# Adversarial Attack on Facial Recognition using Visible Light

Morgan Frearson

Kien Nguyen

*Science & Engineering Department, Queensland University of Technology*
*2 George St, Brisbane City QLD, Australia*
morgan.frearson@connect.qut.edu.au
k.nguyenthanh@qut.edu.au

*Abstract*— The use of deep learning for human identification and object detection is becoming ever more prevalent in the surveillance industry. These systems have been trained to identify human body's or faces with a high degree of accuracy. However, there have been successful attempts to fool these systems with different techniques called adversarial attacks. This paper presents a final report for an adversarial attack using visible light on facial recognition systems. The relevance of this research is to exploit the physical downfalls of deep neural networks. This demonstration of weakness within these systems are in hopes that this research will be used in the future to improve the training models for object recognition. As results were gathered the project objectives were adjusted to fit the outcomes. Because of this the following paper initially explores an adversarial attack using infrared light before readjusting to a visible light attack. A research outline on infrared light and facial recognition are presented within. A detailed analyzation of the current findings and possible future recommendations of the project are presented. The challenges encountered are evaluated and a final solution is delivered. The projects final outcome exhibits the ability to effectively fool recognition systems using light.

This paper is a continuation of the work done in a previous paper 'Building an Adversarial Hat to Fool Facial Recognition' by Morgan Frearson.

GITHUB LINK for LPO - https://github.com/loumor/Adversarial-Attack-LPO-

## I. Introduction

Adversarial attacks aim to move an object's class across the decision boundaries of a DNN causing that object to be misclassified. This form of attack exposes fundamental blind spots in the training algorithms of DNN's [1]. A major adoption of these object detection systems by the surveillance industry (CCTV) has put pressure on the accuracy. Facial identification and human detection are two of the most prevalent and sort after features for a CCTV network.

Specifically, facial recognition systems have demonstrated an accuracy of 99.63% when identifying individuals [2]. To yield this result clean input data with no malicious intention from the individual is processed by the FR system. Several researchers have presented techniques to attack these systems via physical adversaries. Work done by Zhu et al. [3] exploited flaws in FR training polices when they managed to decrease the confidence score of an individual using targeted makeup on the face. However, the most prevalent research towards fooling FR systems has been done using IR light. Zhou et at. [4] and Yamada et at. [5] both published successful papers utilizing IR light to misclassify FR. Their methods exhibited the strongest form of attack for the smallest resources required.

It was established that an IR adversarial attack on FR would be ideal under the project scope. The aim was to create a wearable hat containing LED's to project IR light onto an attacker's face with the intention to misclassify the individual on the FaceNet system. The following paper explores the final design, testing and refinement of this IRH as well as, the LPO system.

A re-evaluation of the adversarial design is explored within, after results from the testing stage produced key challenges on the IRH. The adversarial focus had shifted to a visible light-based attack rather than the originally proposed IR light. The results from this new form of attack proved promising and are discussed within this paper. Key challenges encountered, limitations and procedures of the system are also discussed. The hope remains, that exposing the downfalls of FR in this project will lead to a better understanding that can be made to prevent such attacks in the future.

## II. IR AND FR LITERATURE REVIEW

### A. IR Systems

IR light has a wavelength that causes it to be invisible to human eyes. However, IR light is detected by cameras as their sensors have a wider frequency range [3]. This results in photos and videos being susceptible to IR sabotage when it is present in the field of vision. An attack from IR can distorted images, making it an adversary to FR.

Yamanda et at. [5] invented a wearable device which makes the human face undetectable in capture images. Their type of adversarial attack relied on attaching IR LED's to a set of glasses (Fig. 1). The attack generated an IR noise signal between 800 to 1000 nm. This spectrum was ideal as they found that most digital cameras operate between 200 to 1100 nm, while humans can only see 380 to 780 nm. Because of the perceived noise generated the facial recognition system was unable to detect a face. This research showed an intuitive way to fool FR however, the current reliability of the work could be scrutinized as facial detection and cameras have improved dramatically since 2012.

Rather than facing the IR light toward the camera Zhou et al. [4] took an approach that projected IR light on to the face. The work of Zhou et al. [4] launched two possible types of adversarial attacks on facial recognition systems. Both forms of attack used IR LED's placed inconspicuously in the brim of a hat (Fig. 2). The IR light was projected onto the attacker's face to either *dodge* the recognition system or *impersonate* someone else. Their method used white-box testing so that light spots (perturbations) could be optimized through the loss function based on the variables of; shape, colour, size and brightness. For the *dodge* attack the aim was to increase the loss between the attackers feature vectors and the threshold to classify said features.

For the *impersonate* attack the aim was more detailed. The system initially searched a database to find an adversary (face) that can be impersonated by the attacker (70% chance of success at this stage). The optimizer then used the technique previously talked about to calculate the necessary perturbations, with the goal to bring the attackers feature vectors into the adversaries. The researchers noted the method was not as reliable under circumstances where the attack was moving their face around. Nevertheless, the findings revealed strong results that this form of attack can fool FR.

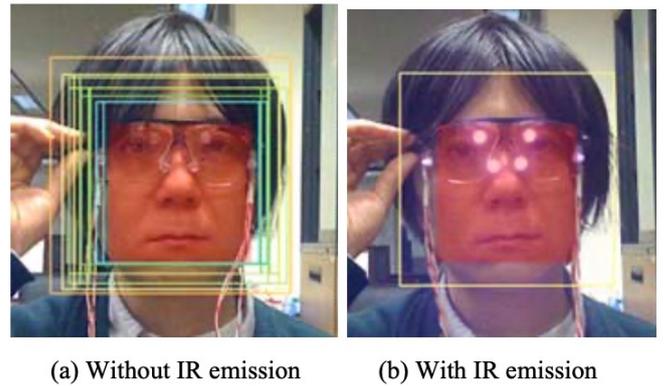

(a) Without IR emission      (b) With IR emission

Fig. 1 Wearable IR LED's glasses also known as 'Privacy Visor'. The lack of squares surrounding the face in (b) denotes a lower confidence level on the system. [3]

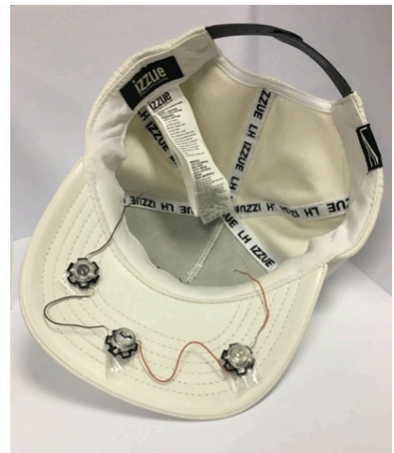

Fig. 2 Wearable IR LED hat. The light is projected onto the face from the three LED's on the brim of the hat. [4]

### B. FR Systems

DNN are large neural networks organized into layers of neurons, corresponding to successive representations of the input data [5]. Each neuron represents an individual computing unit that applies an activation function to the input before passing it onto the next neuron. The architecture of a network relies on weights and biases that characterize strengths of relationships between neurons. A convolutional neural network (CNN) is a type of DNN specific for computer vision tasks. A popular type of CNN for FR is Google's FaceNet system. The system promises excellent results with an accuracy of 99.63% on the LFW dataset [2]. As stated by Schroff et al. [2], *"FaceNet is a system that directly learns a mapping from face images to a compact Euclidean space where distances directly correspond to a measure of face similarity. Once this space has been produced, tasks such as face recognition, verification and clustering can be easily implemented using standard techniques with FaceNet embeddings as feature vectors"*. Once the L2 distance between two faces has been measured the system employs a triplet loss function. This minimizes the distance between a positive anchor while maximizing distances to a negative anchor ( Fig. 3). The aim is to make the squared distance between similar faces smaller and the squared distance between dissimilar faces larger inside the feature space. In other FR systems the loss encourages all faces of one identity to be projected onto a single point in the feature space. Whereas, triplet loss enforces margins between each pair of faces from one identity thus there is stronger discriminability to other identities. Further details on the implementation of the FaceNet system corresponding to the project is addressed in section III. Research Progress.

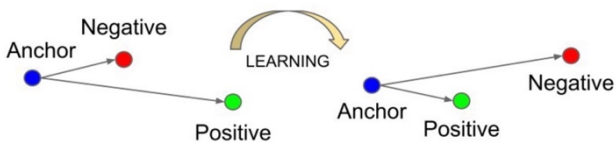

Fig. 3 Triplet Loss minimizes the distance between an anchor and a positive, both of which have the same identity, and maximizes the distance between the anchor and a negative of a different identity.
[2]

*C. One Pixel Attacks*

Su et al. [7] presented research on adversarial attacks under a limited scenario where only one-pixel could be altered on the input image. The attack takes a black-box approach using differential evolution to optimize the one-pixel target position. It does this by generating a set of candidate solutions (children) and comparing them to their parent set, if the children possess higher fitness values they survive over the parents. The result is an x-y coordinate and RGB value of the one-pixel perturbation. The initial aim was un-targeted misclassification of objects, not only was this successful but further findings demonstrated that the location and RGB value of the one-pixel attack could be used misclassify objects into a limited number of targeted classes (Fig. 4). The general conclusion drawn from the paper established that similar class pairs are much easier to perturbed, for example, a cat can easily be classed as a dog but much harder to class as an automobile. This information is valuable as it illustrates that some classes are more robust, indicating their decision boundaries are less likely to overlap. This one-pixel attack is a perfect example that the smallest alteration on an input image can have a dramatic effect on the output classification.

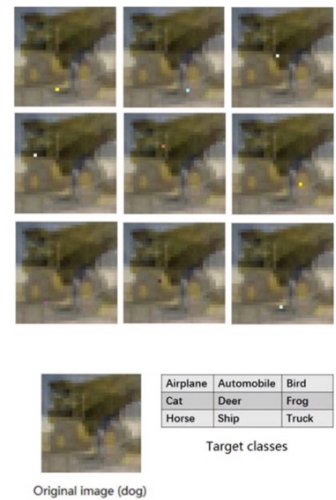

Fig. 4 The dog class can be perturbed to other targeted classes based on the x-y location and RGB value of the pixel. [6]

III. RESEARCH PROGRESS

From the literature it was originally established that an IRH would be built to fool a facial recognition system. The following section discusses the IRH based adversarial attack.

## A. Final Design

Building the IRH was a smooth process with limited issues. There were two adjustments made to the during the construction; current limiting resistor size and added lenses. Due to limitations with available resistor sizes the 0.33 Ω was changed for a 0.47 Ω, this produced V = 1.7 v and I = 1060 mA through each LED [7]. The final LTspice schematic is shown in Figure 5. According to the LED data sheet [8] this lower current ($I_{Recommended}$ = 1400 mA) was still practical except relative luminesce decreased by ~ 5 - 10 %. There was some initial concern with the angle of light emitted from the LED's as light distribution emits at 140 º [8]. To concentrate this a 15 º lens was added to each LED, this improved the brightness and telescoped each light spot. Velcro is incorporated on the brim of the hat to allow for position adjustments of the LEDs. The final design is shown in Figure 6, note that the battery is attached at the back of the hat.

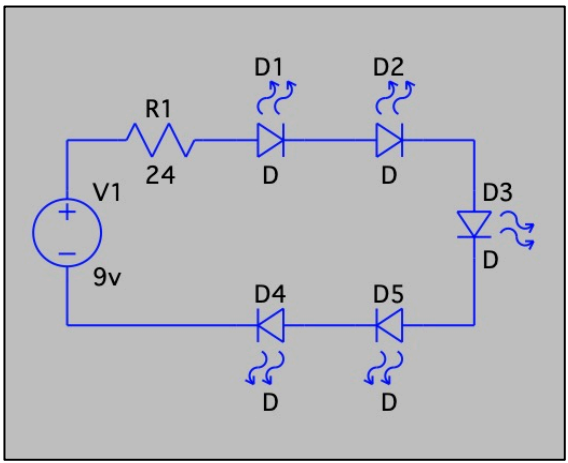

Fig. 5  LT Spice circuit diagram for IRH.

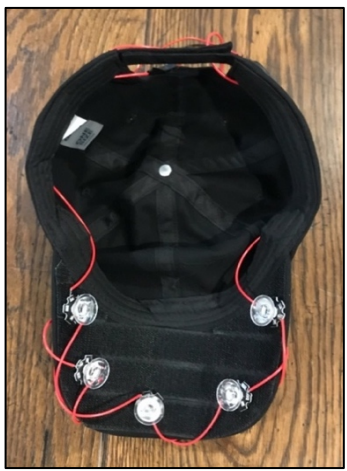

Fig. 6  The IRH final build

Preliminary testing of the IRH produced conflicting results shown in Table I. The LEDs were randomly positioned for the photo (MacBook Pro inbuilt camera) as there was no intention of a specific attack. Very positive results came from the FD system, showing a decrease in confidence by 24.65% ~ IRH [on]. However, for the same photo the FR system increased in confidence by 15.45% ~ IRH [on]. From this, it's possible that the previous assumption of black-box transferability between the FD to FR systems may not be achievable as the results contradict. However, until official testing with the LPO is established the original assumption remains true until proven otherwise.

TABLE I
PRELIMINARY TESTING OF IRH ON FD AND FR

| | IRH Off | IRH On | Confidence Difference |
|---|---|---|---|
| OpenCV FD | Confidence = 92.12% | Confidence = 67.47% | ↓ 24.65% |
| FaceNet FR | Confidence = 77.20% | Confidence = 92.65% | ↑ 15.45% |

There were two observations from this test that could be noted.
- There are limitations on the amount of surrounding light. Testing must be performed in low light environments as the CMOS image sensor used on the MacBook Pro includes an Infrared Cut Filter (ICR) that attenuates infrared wavelengths until light diminishes below a certain level [9].

- The bounding box appears to shrink on both FD and FR systems when IRH [on]. The reason for this is unknown but it is possible that where the light is focused (predominantly forehead area) tightens the bounding box around it.

*B. LPO System*

In order to determine the adversarial positioning of the light spots on the attacker's face, a light perturbation optimizer is designed. The aim of the LPO is to take an initial input photo of the attacker with the IRH off, from here the LPO calculates the optimal positioning for each light spot in order to minimize the confidence score of the detection. After this the attacker positions the LED's on the IRH to match the results from the LPO. To run this system in black-box conditions the attacker assumes no knowledge of internal workings of the system being used (FaceNet FR) and can only access the returned confidence scores. Figure 7 demonstrates how to use the LPO.

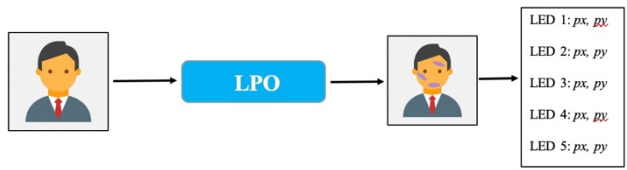

Fig. 7 LPO working diagram. The LPO returns the center [px, py] of each light spot on the face.

The attackers aim is to ultimately lower their confidence score as an untargeted attack on the FaceNet FR system. To do this without the internal knowledge of the system or its specifically trained model is difficult. Instead, an assumption is made that if we can lower the confidence score of detecting a face then it would also deter recognition of a face. For the project two systems are setup; the first is a FaceNet system trained to recognize the attacker, the second is a general OpenCV system trained to detect faces.

The FaceNet system is the FR. This is built in Python by *Aras Futura*. The system is trained on a random selection of 1150 subjects from the LFW dataset and the attacker (me). Due to limitations with the computing environment it was not possible to train on more subjects as processing time increased dramatically.

The OpenCV system is the FD. This is built in Python by *Adrian Rosebrock*. The system uses a pretrained model that detects faces. This FD is the system that the LPO is going to be built on top of.

The idea is to achieve adversarial attack black-box transferability from the FD to FR. The LPO uses the confidence scores of the FD to find the optimal positioning of light spots to render an attack. The attacker then adjusts the LEDs on the IRH to mimic the LPO's results and attempts the attack on the FR system. Figure 8 visualizes this process.

Due to time constraints of the project the variable parameters were kept minimal. On the IRH the [*px,py*] of the light spots is the only adjustable component. In research by Zhou et al. [4], the brightness and size of light spots were variables that could be altered as well. Currently the brightness and light size on the IRH stay constant and the LPO only determines positioning.

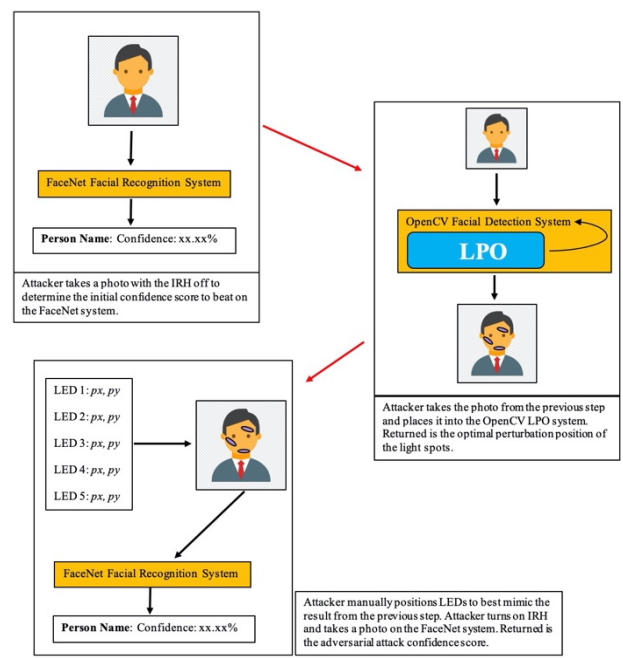

Fig. 8 LPO full process.

For the LPO, a perturbation spot model is devised with three factors to be as close to the real one as possible; colour, shape and opacity. The RGB colour is obtained from a sample photo of the light on the attacker's face. The shape is difficult as light contours to the surface its displayed on, for this an estimated shape from a series of samples is devised. For opacity, estimated guesses are made until the model appears like the sample photo. The final computer-generated version of this is shown in Figure 9. This light spot is developed to be as close to the natural light spot produced by an individual LED on the IRH. The file is saved as an *.png* so during the processing stages there is no need to remove background pixels.

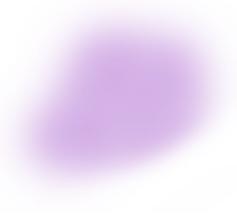

Fig. 9 Individual computer-generated version of the natural light spot on the IRH.

Within the LPO the attackers face is initially cropped using OpenCV tools and resized to fit the frame. The point of this step is to bring all faces to the same size no matter how far away the attacker is standing. The perturbation spot model is then applied to the face. The new photo is fed into the FD and the confidence score is returned. The LPO continues using a simple brute force search to shift the light spot and obtain the confidence score for all possible combinations. This method is similar to what Su et al. [7] performed for the one-pixel attacks. The final step of the LPO is returning the lowest confidence score and corresponding center [*px,py*] of each light spot.

**LPO Directory Tree:**

```
├── detect_faces.py
│            : The main entry point of
  the LPO and the FD system.
│
├── facenet_pytorch_specific
│   │         : Directory containing the
  FR system files.
│   │
│   ├── face_recognition
│   │   │       : Directory containing FR
  functions for recognition processing.
│   │   │
│   │   └──
  face_features_extractor.py
│   │       ├── face_recogniser.py
│   │       └── preprocessing.py
│   ├── images
│   │         : Directory containing all
  images of the 1150 subjects for training.
│   │
│   ├── inference
│   │   │       : Directory containing
  scripts used to train the FR.
│   │   │
│   │       ├── classifier.py
│   │       ├── constants.py
│   │       └── util.py
│   ├── model
│   │   │       : Directory containing the
  trained model for FR.
│   │   │
│   │       └── face_recogniser.pkl
│   ├── requirements.txt
│   │         : File denotes the
  dependencies of the program.
│   │
│   ├── tasks
│   │   └── train.sh
│   │         : Shell script is used to
  retrain the model.
│   │
│   ├── training
│   │   └── train.py
│   │         : Training functions are set.
│   │
│   └── util
│       │     : Directory containing
  Facenet functions for FR.
│           ├── align_mtcnn.py
│           ├── collect_face_images.py
│           ├── generate_embeddings.py
│           └── tsne_visualization.py
├── images
│   │         : Directory containing
  input images to start the LPO. The light
```

```
|   |                    spot .png file is also
stored here.
|   |   ├── front.jpg
|   |   ├── left.jpg
|   |   ├── lightspot.png
|   |   └── right.jpg
├── output_images
|   |               : Directory containing
output images of the LPO.
|   |   ├── front_Final_FD.jpg
|   |   |           : Attacker's front being
tested on the FD ~ IRH [on].
|   |   ├── front_IRHON_Final_FR.jpg
|   |   |           : Attacker's front being
tested on the FR ~ IRH [on].
|   |   ├── front_Inital_FD.jpg
|   |   |           : Attacker's front being
tested on the FD ~ IRH [off].
|   |   ├── front_Initial_FR.jpg
|   |   |           : Attacker's front being
tested on the FR ~ IRH [off].
|   |   ├── front_LPO.jpg
|   |   |           : Attacker's front LPO
suggestion ~ IRH [off].
|   |   ├── left_Final_FD.jpg
|   |   |           : Attacker's left being
tested on the FD ~ IRH [on].
|   |   ├── left_IRHON_Final_FR.jpg
|   |   |           : Attacker's left being
tested on the FR ~ IRH [on].
|   |   ├── left_Inital_FD.jpg
|   |   |           : Attacker's left being
tested on the FD ~ IRH [off].
|   |   ├── left_Initial_FR.jpg
|   |   |           : Attacker's left being
tested on the FR ~ IRH [off].
|   |   ├── left_LPO.jpg
|   |   |           : Attacker's left LPO
suggestion ~ IRH [off].
|   |   ├── right_Final_FD.jpg
|   |   |           : Attacker's right being
tested on the FD ~ IRH [on].
|   |   ├── right_IRHON_Final_FR.jpg
|   |   |           : Attacker's right being
tested on the FR ~ IRH [on].
|   |   ├── right_Inital_FD.jpg
|   |   |           : Attacker's right being
tested on the FD ~ IRH [off].
|   |   ├── right_Initial_FR.jpg
|   |   |           : Attacker's right being
tested on the FR ~ IRH [off].
|   |   └── right_LPO.jpg
|   |               : Attacker's right LPO
suggestion ~ IRH [off].
|   |
├── res10_300x300_ssd_iter_140000.caffemodel
|                   : File containing the trained
model for FD.
└── run.sh
                : Shell script used to run the entire
program.
```

**run.sh Shell Script:**

Figure 10 below is a screenshot of the shell script required to run the LPO program. To run the script the image path for the initial photos ~ IRH [off] of the attacker's face must be included.

The variable *PIX_INCR* allows the user to set the number of pixels they would like to shift the light spot for every iteration. For example; setting *PIX_INCR* at 50, would have the LPO test the light spot at 50px increments, 0, 50, 100, 150…. The smaller *PIX_INCR* the more accurate but time consuming the LPO becomes. Be cautions of your operating system's available stack space when deciding on the *PIX_INCR* size. Because the black-box testing method nested *for* loops are used inside a single function, this can start to use up stack space fast if *PIX_INCR* < 50 for the average computer.

Fig. 10  run.sh Sell Script to run the LPO program.

The program runs the FR system for the front, left then right initial photos ~ IRH [off] to obtain the initial confidences scores. This saves the *.jpg* files in the *../output_images* folder. From here the LPO system is run inside the *detect_faces.py* file. The user is prompted to retake the photos with ~ IRH [on] after the LPO has produced its suggestion. Finally, the program reruns the FR system for the front, left, then right photos ~ IRH [on] to obtain the final confidences scores.

**Important Functions of detect_faces.py (LPO system):**

Function – `initialConfidence`
   Inputs – `imagesFLR, net`
   Returns – `face_locationFLR, imageWithConfidenceFLR, image_dimesnsionsFLR`
   Description – This function takes in the pretrained model and an array containing the front, left and right initial images of the attacker's face. Processing is done to determine the location and confidence of the face. Returned is an array containing the location of each face [$x_1, y_1, x_2, y_2$], image with bounding boxes and dimensions of each image [$p_x, p_y$].

Function – `cropImageFLR`
   Inputs – `face_locationFLR, imagesFLR`
   Returns – `crop_imgFLR`
   Description – This function takes in an array containing the location of each face and the original images. Using the location of the face the original image is cropped to fit this. Returned is an array of the cropped faces.

Function – `lightSpotModel`
   Inputs – `image_dimensionsFLR, face_locationFLR, imagesFLR`
   Returns – `lightspotFLR`
   Description – This function takes in an array containing the image dimensions, face location and original image of each face. Here the generated light spot model (Fig. 9) is rescaled based on the current face size compared to the original face size that the light spot model was composed from. Returned is an array of the correctly scaled light spot, appropriate to each face.

Function – `LPO_Front`
   Inputs – `crop_imgFLR, lightspotFLR, face_locationFLR, image_dimensionsFLR, imageFLR, pixelIncrement, net`
   Returns – `lpoFrontImage`
   Description – This function is where the LPO brute force search is conducted. The inputs are all specific to the front facing image of the attacker. This is because the front of the IRH has three LEDs, thus the algorithm must account for three light spots. Nested *for* loops are used to increment the *x* and *y* position of the light spot to iterate across all possible positions. Returned is the front image with the recommended light spot positioning for the attacker to achieve the lowest confidence score.

Function – `LPO_LR`
   Inputs – `crop_imgFLR, lightspotFLR, face_locationFLR, image_dimensionsFLR, imageFLR, pixelIncrement, net`
   Returns – `lpoLeftImage(repeated for right)`
   Description – Same as the `LPO_Front` description however, only one light spot is required as the IRH has only one LED on the left and right sides.

Function – `faceDetection`
Inputs – `image, net`
Returns – `image, confidence`
Description – This function takes in an input image and the pretrained model. Using OpenCV the face is detected, and the bounding box is drawn. The detected image and confidence score are returned.

As mentioned prior the *PIX_INCR* allows the user to adjust the number of pixels the LPO shifts upon each iteration of the search. This can vastly affect the time taken to run the program due to the time complexity generated from the nested *for* loops in the `LPO_Front` function. In an attempt to speed this up, the python package *threading* was imported, and a multithreading implementation of the function was run. Unfortunately, this parallelization slowed the program down further due to two factors; there is a shared resource that costs on the context switching and there is an I/O bound bottleneck issue when opening the image resource on each iteration. Because of this the multithreading implementation was removed from the program.

**Operational Testing of LPO and IRH**

To start the LPO navigate to directory and type the command shown in Figure 11. Remember to include the image paths to the initial front, left and right images as well as your desired pixel increment size.

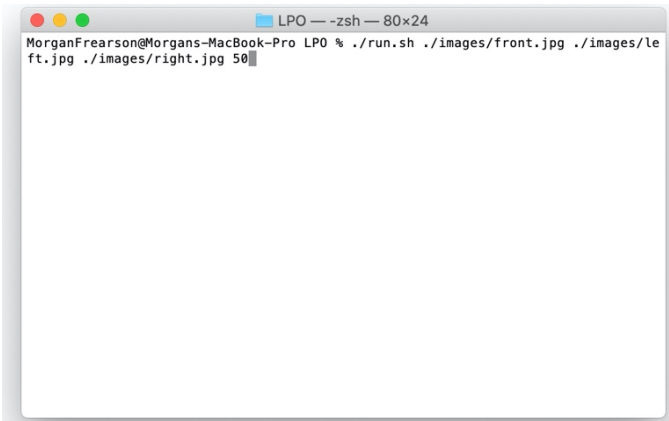

The photos being used for this test are shown in Figure 12.

Fig. 11 Command to run the LPO.

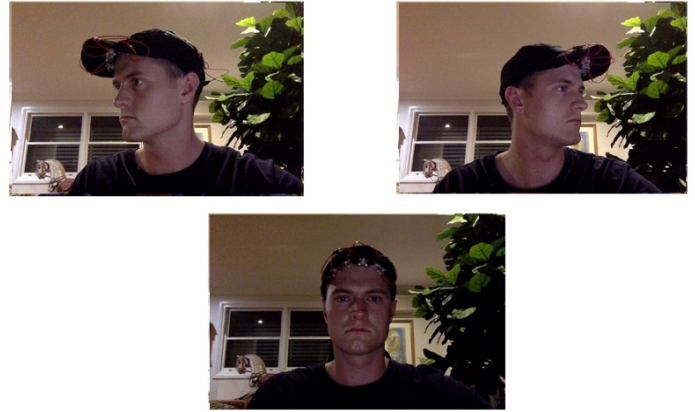

Fig. 12 Initial test images (Left, Right, Front).

Once the LPO has run you will get a screen shown in Figure 13. The LPO has saved its recommendation including the initial confidence scores of the FR and FD system to the *./output_images* folder (these images are shown below).

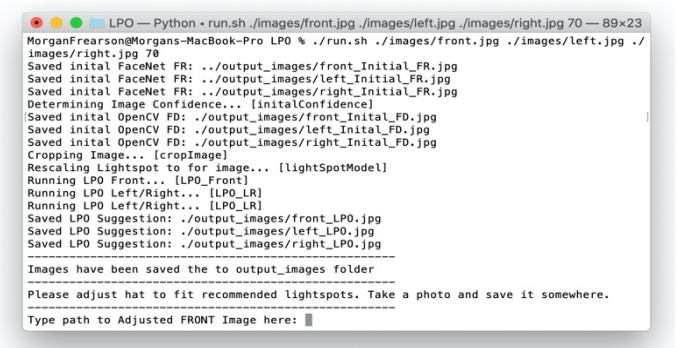

Fig. 13 Screen after running the LPO.

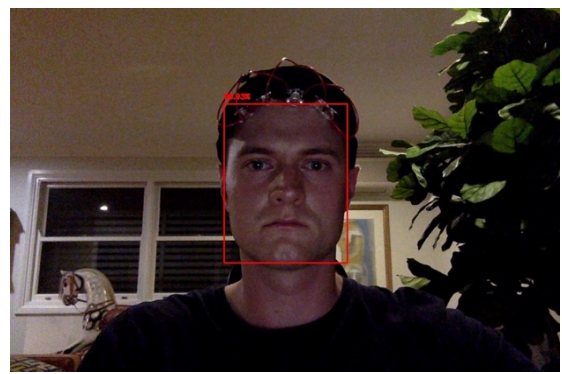

Fig. 14 Front FD Confidence (99.93%).

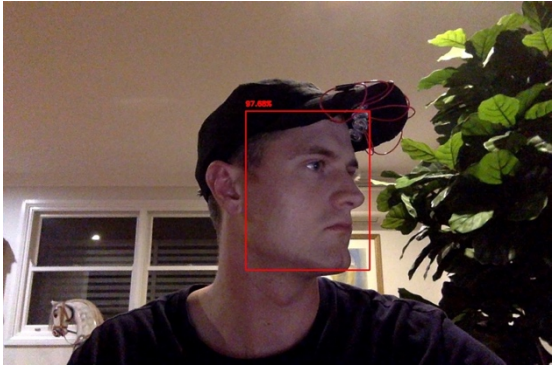

Fig. 15  Right FD Confidence (97.68%)

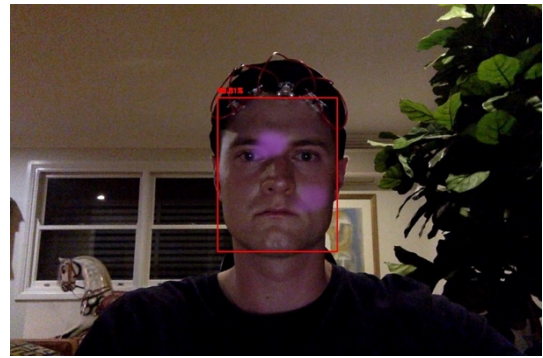

Fig. 19  Front LPO Recommendation (99.81%)

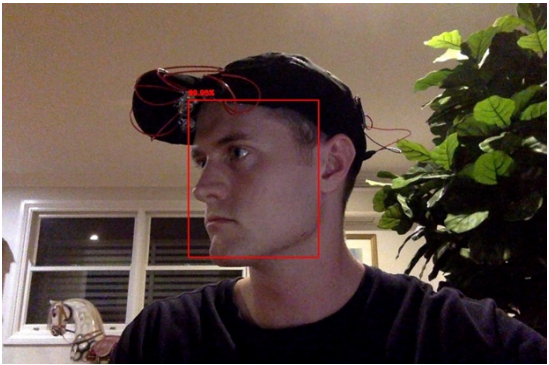

Fig. 16  Left FD Confidence (99.95%)

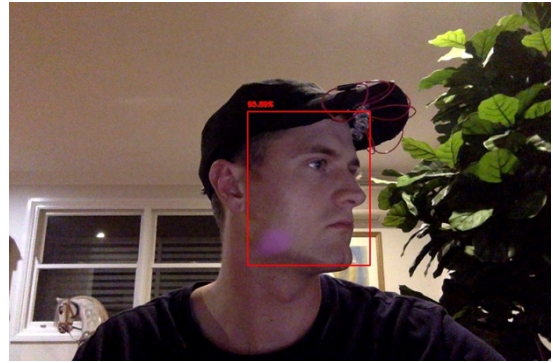

Fig. 20  Right LPO Recommendation (95.89%)

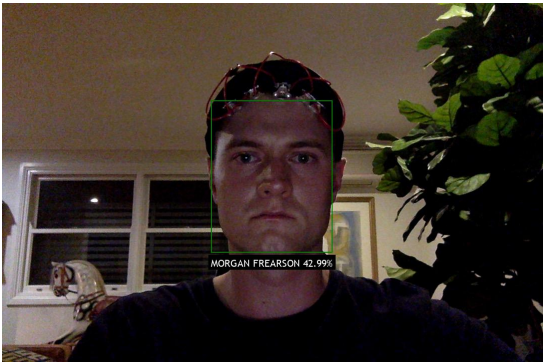

Fig. 17  Front FR Confidence (42.99%)

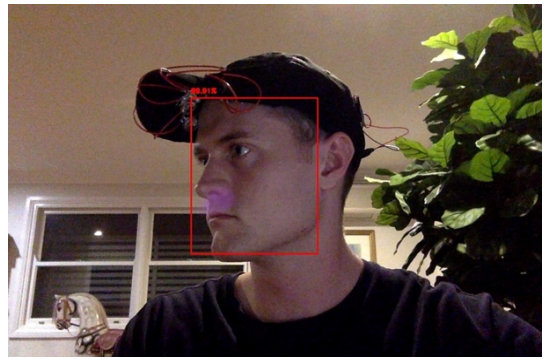

Fig. 21  Left LPO Recommendation (99.91%)

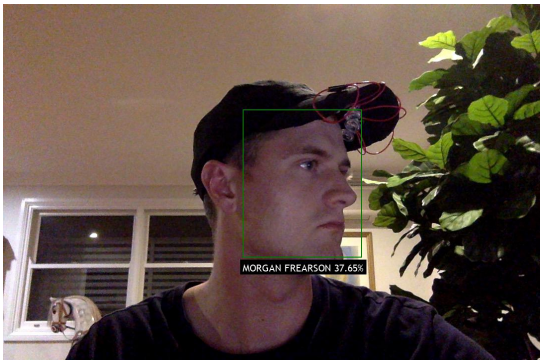

Fig. 18  Left FR Confidence (24.28%)

After examining the LPOs recommendations the user can now adjust the IRH to mimic the light spots to the best of their ability. As depicted in Figure 22, the user types the image path of the new updated photos ~ IRH [on]. The resulting images for this are shown below.

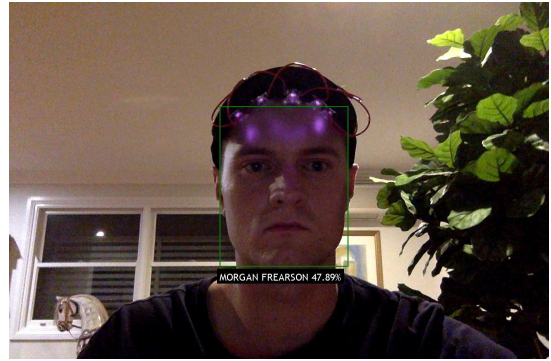

Fig. 22  Screen to type image path IRH~[on].

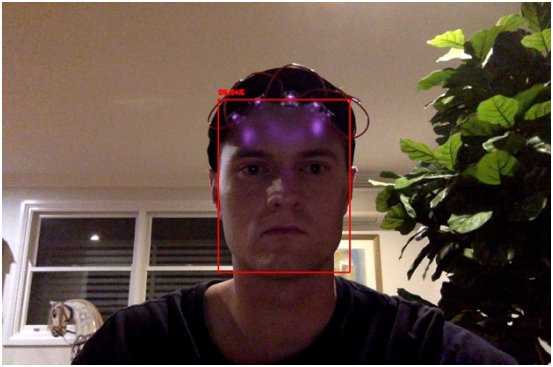

Fig. 23  Front FD ~IRH [on] Confidence (99.64%)

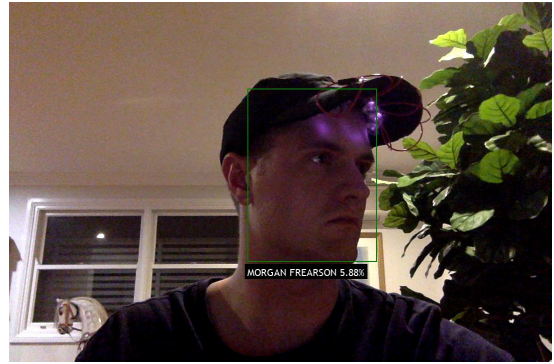

Fig. 26  Front FR ~IRH [on] Confidence (47.89%)

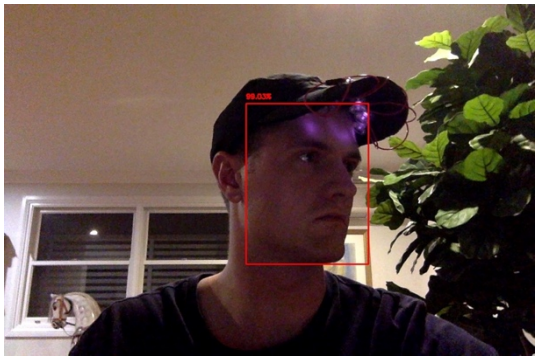

Fig. 24  Right FD ~IRH [on] Confidence (99.03%)

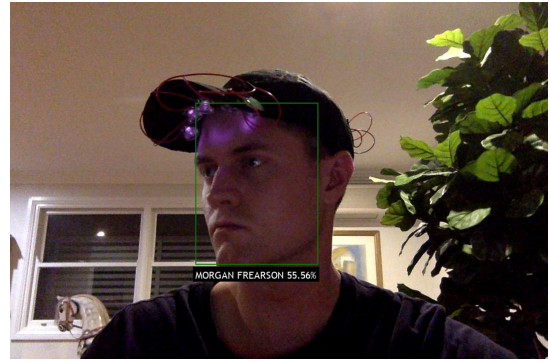

Fig. 27  Right FR ~IRH [on] Confidence (5.88%)

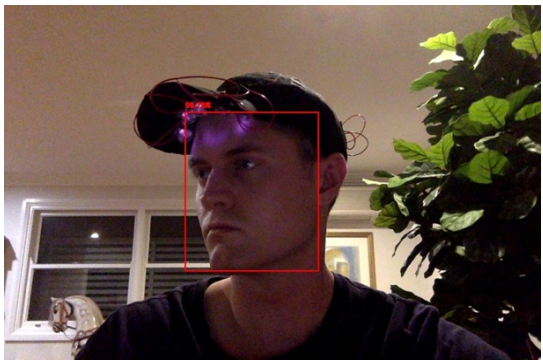

Fig. 25  Left FD ~IRH [on] Confidence (99.75%)

Fig. 28  Left FR ~IRH [on] Confidence (55.56%)

When the LPO has finished a set of general statistics are printed out to help with debugging or data observation. This is shown in Figure 29.

Fig. 29 Final screen after running the LPO.

This initial test of the LPO proves that the black box method is successful at producing recommendations for the lowest confidence positioning of the light spots. However, during this testing several obstacles became evident which are potential damaging on the ability to obtain accurate results.

Firstly, the IRH has limited range on how far the light can be moved on the face. This is demonstrated in Figure 23-25 where you can see that the light spots are concentrated on the forehead. This is because of the IR lights weak ability to be observed under luminescent situations, as the surrounding light becomes brighter the IR light becomes dimmer. In dim conditions the light spots can be observed on the nose but, even in complete darkness the IRH struggles to reach the chin (Fig. 30).

Secondly, light is extremely hard to model and the LPO only uses a single standard spot type shown in Figure 9. Because of the way light is meant to mould to the depth of a face or the curvature the LPO doesn't produce accurate results. The ability to accurately model light on the face is well out of the timeframe of this project.

Finally, low light testing which was observed as a limitation earlier adds unnecessary constraints to the testing environment. Having to adjust the surrounding light is subjective, each testing period this is done as an estimate and is not consistent.

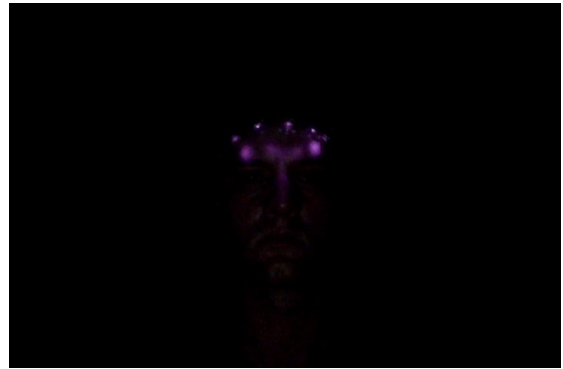

Fig. 30 IRH in complete dark conditions struggling to reach past the nose.

### C. Refinement

The IRH works well but is limited to only the top half of the face which puts a large constraint on testing. Furthermore, the light of the IRH is not accurately represented on the LPO due to the use of a single standard light spot (Fig. 9). The ability to correctly model light on the face is out of the timeframe of this project and is not the main focus. For these reasons a slight change of the project objectives must be made. The following section explores a new type of adversarial attack that still involves light and utilizes the LPO.

To gain accurate results from the LPO the new adversarial attack must include a smaller more stable-shaped light spot, so that the model is as close to the realistic light spot as possible. Having a smaller circular light spot would diminish the problem of the LPO not being able to adapt to the depth and curvature of a face as a real light spot would. Looking at the One-Pixel attack [7] shows that a small change on an image can alter the classification class. This is what I am hoping to achieve with this new form of adversarial attack, by having a refined light spot act in the way that the removal of pixel does in Su et al. [7] paper. Considering all these aspects the appropriate way to generate this new light-based adversarial attack is with a laser pointer.

The LZP acts as the new perturbation on the face instead of using the IRH. This new attack type is now an attempt to move the work done by Su et al. [7] into the physical realm. The main benefit from using the LZP is that it can reach all parts of the face without any variation in luminescence. To keep the shape consistent a single point/distance from the user will be set before testing, this is demonstrated in Figure 31. High caution will be taken when pointing the LZP in the direction of the face, a safe padding distance will be established around the eye area to avoid any accidents. The LZP being used is 100mW 532nm (green) and requires two AAA batteries. Figure 32 shows what the LZP looks like when used on the face.

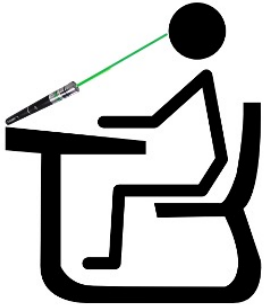

Fig. 31   LZP locked at a single consistent distance from the face to keep the luminescence and shape similar per test.

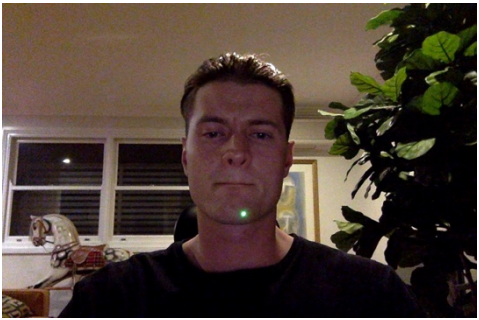

Fig. 32   The real LZP example acting on the face.

For the LPO, a new perturbation spot model is devised with three factors to be as close to the real one as possible; colour, shape and opacity. The RGB colour is obtained from a sample photo of the LZP on the attacker's face. The shape of the LZP is much more consistent than that of the IRH. The light still contours to the surface its displayed on however, because the spot is smaller the contouring is less noticeable, and the shape tends to remain circular. For opacity, estimated guesses are made until the model appears like the sample photo. The final computer-generated version of this is shown in Figure 33. This light spot is developed to be as close to the natural light spot produced by a single LZP.

The file is saved as an *.png* so during the processing stages there is no need to remove background pixels.

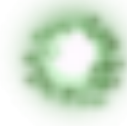

Fig. 33   Individual computer-generated version of the natural light spot LZP.

Figure 34 compares what the real LZP looks like verses the computer-generated version that will be used in the LPO. The major downside of using the LZP instead of the IRH is the attack is no longer invisible to the human eye. However, this is a tradeoff because the LZP is easier to setup, as such the time taken to orchestrate an attack on a CNN is the upside. In Su et al. [7] paper the input images were unified to a 227 x 227-pixel resolution. Dragging the input pixels lower means that in proportion the removal of one-pixel has a bigger effect on the image. Therefore, this is likely to have a proportional effect on the classification results. This strategy will be tested under this new attack with the LZP to see if decreasing the quality of the images improves the misclassification and the adversarial abilities of the attack.

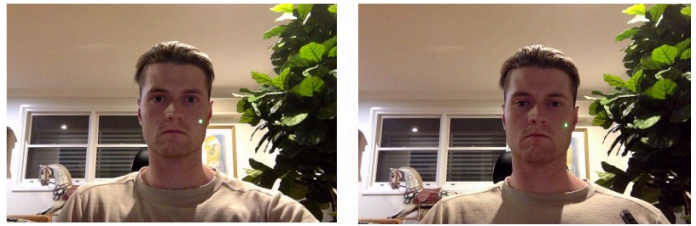

Fig. 34   Computer generated LZP (Left) compared to the real LZP (Right). More realistic than the IRH light spot.

IV. FINAL RESULTS

A. Updated LPO

A small adjustment to the LPO had to be made so that it was compatible with the LZP attack. In the original IRH the front image produced three light spot perturbations on the face, as the LZP attack only uses a single light spot the LPO was modified to match this. This made the LPO more efficient as there were less nested *for loops* within the code. The

LPO continues to take photos and provide recommendation on the front, left and right images of the face. The LZP attack is limited to a single light spot per test, so testing the front, left and right at the same time is not possible in the physical sense.

## B. Testing and Results

To maintain a consistent testing environment the location and distance from the camera were kept the same. The LPO was set so that *PIX_INCR* = 20, this was a reasonable range that balanced computing time and black-box depth of analysis. Upon testing it was noticed that the sounding light (LUX levels on the testing image) would vary based on if it was day or night (standard house lights were used at night). On the results tables below the column 'Testing Environment' denotes if a test was taken during the day or at night. To limit the environment effects on the confidence score, random plain colour t-shirts were worn for each test. CNN's are insignificantly affected by simplistic monochromatic clothing [10]. In some testing scenarios a standard hat was worn. In the 'Testing Environment' column a 'H' denotes a test where a hat was worn. Wearing a hat is a good indicator if the LZP attack works under adjusted circumstances by looking at the *FR Diff (%)* and *FD Diff (%)* columns.

TABLE II
FRONT LZP ATTACK TESTING RESULTS. 'D' = DAY, 'N' = NIGHT, 'H' = HAT.

| FRONT | FD Initial % | FD Final % | FD Diff | FD Diff (%) | FR Initial % | FR Final % | FR Diff | FR Diff (%) | Testing Environment |
|---|---|---|---|---|---|---|---|---|---|
| Test 1 | 99.87 | 99.83 | ↓ 0.04 | 0.04% | 71.79 | 59.05 | ↓ 12.74 | 17.74% | D |
| Test 2 | 99.97 | 99.84 | ↓ 0.13 | 0.13% | 59.87 | 43.78 | ↓ 16.09 | 26.87% | DH |
| Test 3 | 99.84 | 90.81 | ↓ 9.03 | 9.04% | 64.54 | 35.07 | ↓ 29.47 | 45.66% | N |
| Test 4 | 81.82 | 97.61 | ↑ 15.79 | 19.3% | 30.11 | 24.63 | ↓ 5.48 | 18.2% | NH |
| Test 5 | 97.57 | 95.22 | ↓ 2.35 | 2.41% | 35.03 | 39.66 | ↑ 4.63 | 13.22% | N |
| Test 6 | 99.91 | 95.77 | ↓ 4.14 | 4.14% | 38.4 | 77.64 | ↑ 39.24 | 102.19% | N |
| Test 7 | 97.95 | 92.26 | ↓ 5.69 | 5.81% | 82.15 | 70.69 | ↓ 11.46 | 13.95% | D |
| Test 8 | 99.73 | 99.97 | ↑ 0.24 | 0.24% | 82.54 | 66.42 | ↓ 16.12 | 19.53% | D |
| Test 9 | 99.92 | 99.71 | ↓ 0.21 | 0.27% | 68.36 | 59.87 | ↓ 8.49 | 12.42% | D |
| Test 10 | 97.58 | 91.3 | ↓ 6.28 | 6.44% | 58.3 | 39.21 | ↓ 19.09 | 32.74% | D |

TABLE III
LEFT LZP ATTACK TESTING RESULTS. 'D' = DAY, 'N' = NIGHT, 'H' = HAT.

| LEFT | FD Initial % | FD Final % | FD Diff | FD Diff (%) | FR Initial % | FR Final % | FR Diff | FR Diff (%) | Testing Environment |
|---|---|---|---|---|---|---|---|---|---|
| Test 1 | 84.43 | 99.99 | ↑ 15.56 | 18.49% | 55.81 | 54.37 | ↓ 1.44 | 2.58% | D |
| Test 2 | 98.79 | 78.54 | ↓ 20.25 | 20.5% | 42.37 | 31.90 | ↓ 10.47 | 24.71% | DH |
| Test 3 | 96.36 | 84.33 | ↓ 12.07 | 12.48% | 47.43 | 31.9 | ↓ 15.53 | 32.74% | N |
| Test 4 | 84.34 | 99.92 | ↑ 15.58 | 18.47% | 43.61 | 44.06 | ↑ 0.45 | 1.03% | NH |
| Test 5 | 98.34 | 94.88 | ↓ 3.46 | 3.52% | 42.24 | 56.25 | ↑ 14.01 | 33.17% | N |
| Test 6 | 97.93 | 83.27 | ↓ 14.66 | 14.97% | 12.25 | 40.05 | ↑ 27.8 | 226.94% | N |
| Test 7 | 80.22 | 96.36 | ↑ 16.14 | 20.12% | 67.23 | 59.03 | ↓ 8.2 | 12.2% | D |
| Test 8 | 81.29 | 99.08 | ↑ 17.79 | 21.88% | 72.96 | 54.86 | ↓ 18.1 | 24.81% | D |
| Test 9 | 80.63 | 87.91 | ↑ 7.28 | 9.03% | 50.49 | 16.05 | ↓ 34.44 | 68.21% | D |
| Test 10 | 95.53 | 99.67 | ↑ 4.14 | 4.33% | 58.14 | 56.09 | ↓ 2.05 | 3.53% | D |

TABLE IV
RIGHT LZP ATTACK TESTING RESULTS. 'D' = DAY, 'N' = NIGHT, 'H' = HAT.

| RIGHT | FD Initial % | FD Final % | FD Diff | FD Diff (%) | FR Initial % | FR Final % | FR Diff | FR Diff (%) | Testing Environment |
|---|---|---|---|---|---|---|---|---|---|
| Test 1 | 82.54 | 97.73 | ↓ 15.19 | 15.54% | 47.52 | 35.33 | ↓ 12.19 | 25.65% | D |
| Test 2 | 93.51 | 85.9 | ↑ 7.61 | 8.14% | 19.7 | 31.84 | ↑ 12.14 | 61.62% | DH |
| Test 3 | 73.4 | 73.76 | ↑ 0.36 | 0.49% | 57.12 | 9.5 | ↓ 47.62 | 83.37% | N |
| Test 4 | 99.14 | 99.7 | ↑ 0.56 | 0.56% | 36.09 | 53.58 | ↑ 17.49 | 48.46% | NH |
| Test 5 | 70.89 | 68.42 | ↓ 2.47 | 3.48% | 52.14 | 25.67 | ↓ 26.47 | 50.77% | N |
| Test 6 | 98.71 | 71.92 | ↓ 26.79 | 27.14% | 8.28 | 9.68 | ↑ 1.4 | 16.91% | N |
| Test 7 | 85.85 | 76.6 | ↓ 9.25 | 10.77% | 26.82 | 27.39 | ↑ 0.57 | 2.13% | D |
| Test 8 | 98.13 | 99.03 | ↑ 0.9 | 0.92% | 59.40 | 35.81 | ↓ 23.59 | 39.71% | D |
| Test 9 | 98.63 | 99.57 | ↑ 0.94 | 0.95% | 40.65 | 32.47 | ↓ 8.18 | 20.12% | D |
| Test 10 | 98.51 | 95.67 | ↑ 2.84 | 2.88% | 41.17 | 31.44 | ↓ 9.73 | 23.63% | D |

An attempt was made to replicate the work done in Su et al. [7] paper and reduce the pixel resolution of the test image. This was unsuccessful in having any effect on the confidence score of the LZP attack. As the LZP is a physical perturbation the reduction in pixel size proportionally reduces the laser spot in the image. Su et al. [7] reduced the pixel size of their test images and then removed the pixel during the postprocessing stage. In theory, overlaying the LZP perturbation on top of the reduced pixel image would cause a reduction in confidence score but this would no longer constitute a 'physical adversarial attack'.

## V. DISCUSSION

Overall the results of the LZP attack are encouraging in its ability to reduce the confidence score on an FR system. On '*Front*' facing scenarios 80% of the tests proved positive in lowering the confidence score using the LZP. '*Left*' scenarios had a 70% success rate while '*Right*' scenarios incurred a 60% success rate. The FD system did not experience as strong results from the LZP attack with; '*Front*' = 80% success, '*Left*' = 40% success, '*Right*' = 40% success.

Firstly, it is worth noting that when the confidence score increased on the FR system the FD system typically resulted in an opposite directional change across the '*Front*', '*Left*' and '*Right*' scenarios. This was also the case when the FD system increased in confidence, the FR system resulted in the opposite. For example, Test 5 & 6 of '*Front*', were the only tests that the confidence score increased on the FR system but for the same tests on the FD system the confidence resulted in a decrease. This discretion between the direction of the confidence score for the FD and FR systems are seen in the majority of the tests across each scenario. The most plausible reason for this occurrence relates to how the FR and FD models were constructed. The FD systems relies on a pretrained model specifically trained to detect faces. Whereas, the FR model was trained based on images of the attacker's face (myself). Looking at the initial FD confidence scores the model detects the face across the tests with a confidence between 70.89 – 99.99%, higher on average than the 19.70 – 82.54% seen with the FR system. This is understandable as the pretrained FD model has a much wider decision boundary making it easier to classify a face rather than a specific individual. Because of this wider decision boundary, the LZP attack doesn't have as much influence on facial features due to it being limited by the single laser spot. On the FR system adding a perturbation (LZP) to a recognizable facial feature would have greater effect on the confidence score. Due to this, the original idea behind the LPO having black-box transferability from the FD to FR appears to be unsupportable. However, the LPO's recommendation still reduced the confidence of the FR system even though the LZP spot is determined by the FD system.

On both the FD and FR systems the '*Front*' scenario returned the most promising results for the LZP attack. Due to the nature of the attack the LZP was manually positioned (Fig. 31). Attempting to position the LZP in '*Left*' and '*Right*' scenarios proved difficult as the angle the attacker is facing is generally away. '*Front*' was more accurate and this consistency is seen in the results (Table II). Figure 35 illustrates the images feed into and out of the LPO for 'Test 3' as an example. Comparing the LPO recommendations to each of their corresponding scenarios (*Front, Left, Right*) shows that the for both the LZP spot and photo consistency the '*Front*' is visually more reliable. This bodes well for the potential of the LZP attack as it is possible to assume that with fine tuning of the testing setup the '*Left*' and '*Right*' attacks could reach the 80% success range (similar to '*Front*').

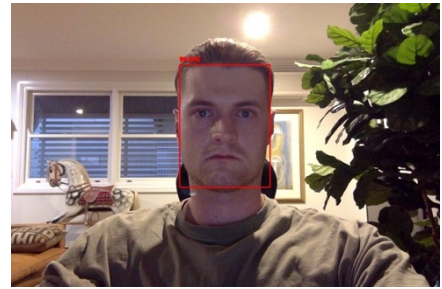

Front Initial FD

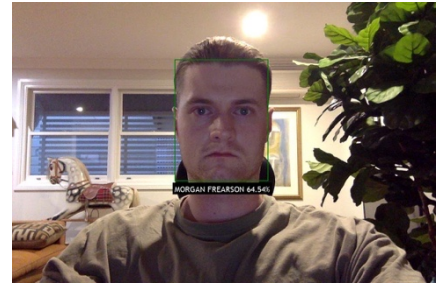

Front Initial FR

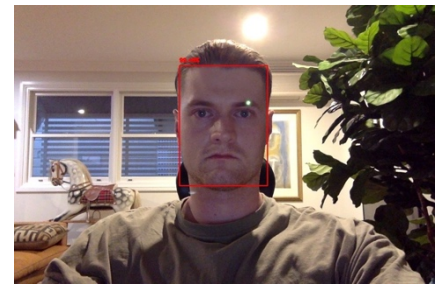

Front LPO

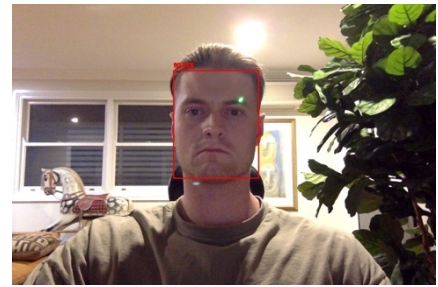

Front Final FD

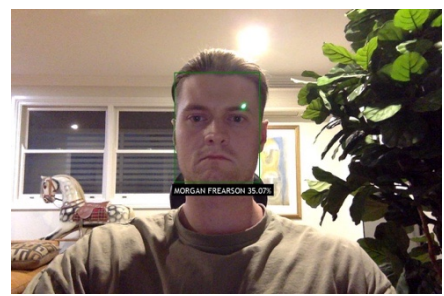

Front Final FR

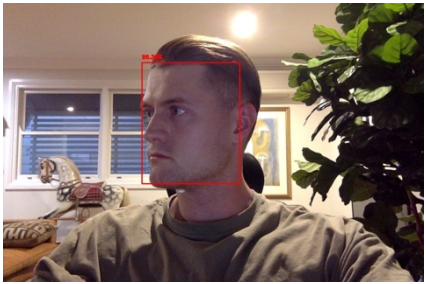

Left Initial FD

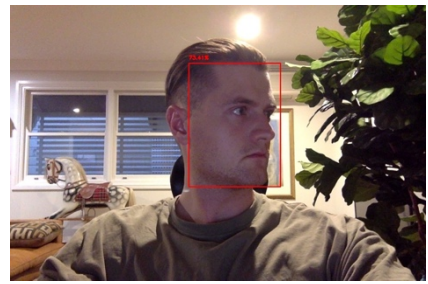

Right Initial FD

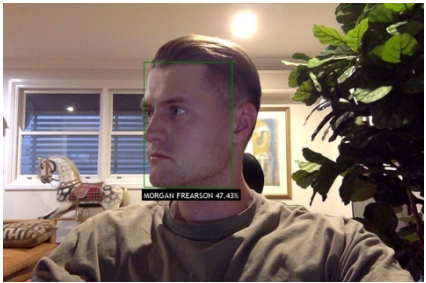

Left Initial FR

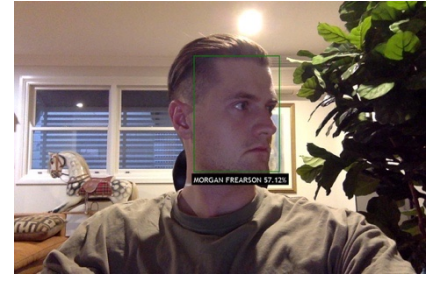

Right Initial FR

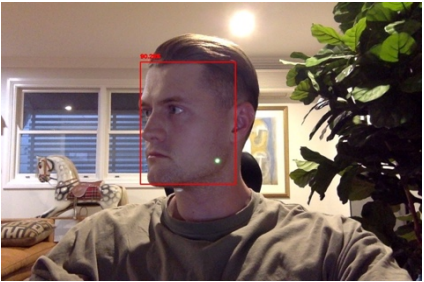

Left LPO

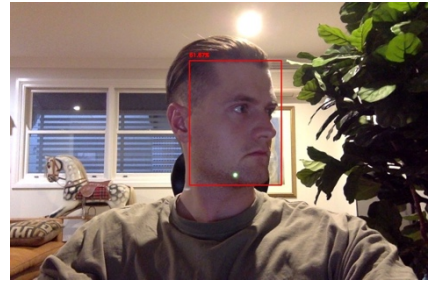

Right LPO

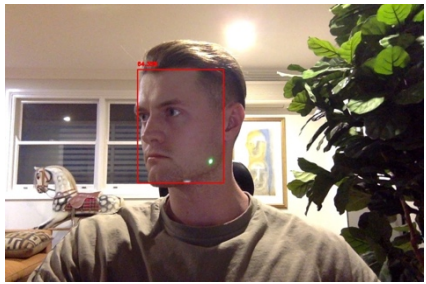

Left Final FD

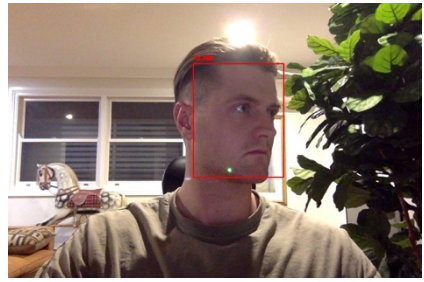

Right Final FD

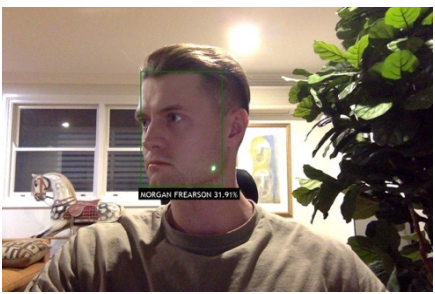

Left Final FR

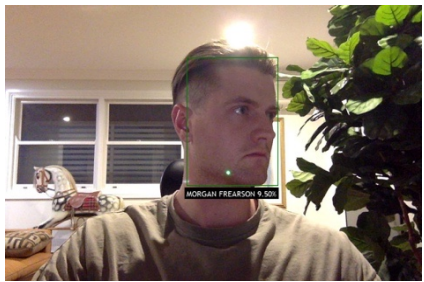

Right Final FR

Fig. 35 Test 3 (Example of testing for the LZP attack).

Even with the LZP attack as simplistic as it is the results reveal an average confidence drop of 28.53% across the successful tests (FR). This is an impressive drop considering the small portion of the face that is perturbated by the LZP. The rate of success for the LZP attack on the FR is also dependent of the number of images that the model is trained on in respect to the attacker. The more images a system is trained on the tighter/stronger the decision boundaries will be [10]. In the instance of the FR system it was trained using the LFW database in conjunction with 24 images of the attacker's face (myself). This was judged as a reasonable number of samples as the LFW database also only ranged between 2 – 30 images per subject. Looking at the '*FR Initial (%)*' in the results tables there is a noticeable discrepancy in the confidence of recognition. The swing in variation is due to the number of samples that the model was trained on (24). This is not necessarily a downfall of the experimental setup but rather just a side effect of using a smaller sample size.

When averaging out the confidence score for failed tests (confidence increased) the result is larger at 56.19% (FR). This is understandable as percentage-based values will have more room to move when increasing (0 - ∞%) rather than the limited 0 – 100% that can be obtained on decreasing values. These failed tests are more than likely due to the black-box transferability not being reliable. If the LPO was built on the FR system, the chances of an increase in confidence would be less likely.

Changes in the attackers positioning can also have an effect on the confidence score. If the attacker changes angle or distance from their '*Initial*' image this can become an advantage or disadvantage to the resulting confidence score comparison ('*Final*' image). This is why the testing was performed in the same environment keeping the alterations between tests at a minimum. Figure 36 shows a how slight movement from the '*Initial*' image can affect the confidence of the '*Final*' image even without the LZP attack. This is an indication of the sensitivity that was present during the testing stage. The final 10 tests that are presented were chosen from a pool of 60 tests. These 10 were chosen because of their minimal positioning change between the '*Initial'* and '*Final*' images in an attempt to keep consistency. Several tests were also unusable due to the LPO making the LZP suggestion over/around the eyes, these were unsafe to perform.

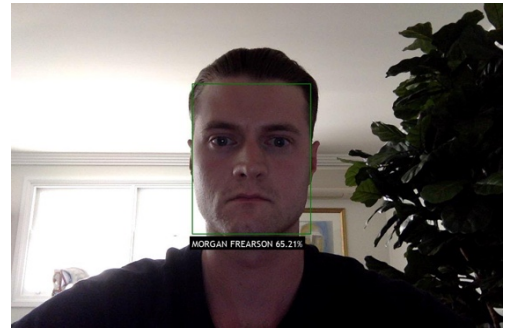
Front Initial

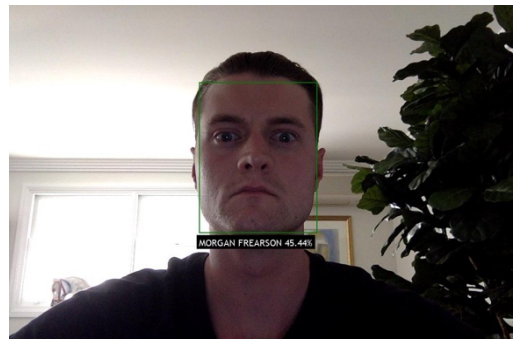
Front Final (Slight Movement Confidence Decrease 19.77%)

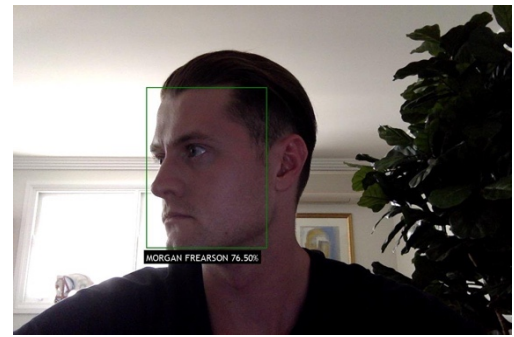
Left Initial

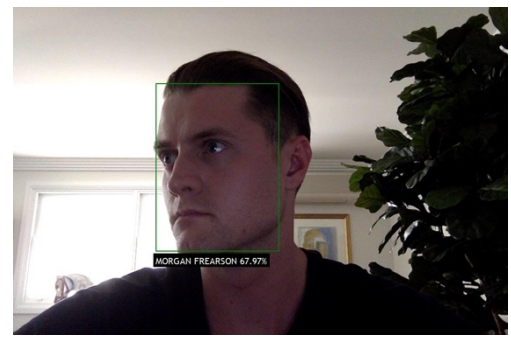
Left Final (Slight Movement Confidence Decrease 8.53%)

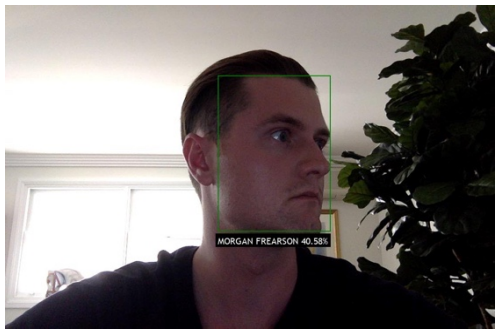
Right Initial

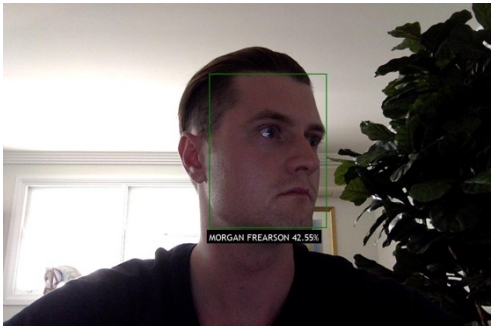
Right Final (Slight Movement Confidence Decrease 1.97%)

Fig. 36 Slight changes in facial positioning can affect confidence score.

VI. CONCLUSION

The research undertaken had the objective to use a light-based device that can fool an FR system. Creating an adversarial attack by adding physical light perturbations to an attacker's face was the starting point of the design. An IRH was constructed but during the testing and refinement stage it was found to have several complications. Because of this a new LZP attack was devised. This device used visible light to implement an adversarial attack similar to Su et al. [7] One-Pixel attack. To optimize the LZP's adversarial ability the LPO system was constructed with black-box testing.

Overall, the LZP attack saw positive results in its ability to reduce the confidence score of an FR system. However, the LZP demonstrated conflicting and negative results on the FD system. Because of this the objective of black-box transferability from the FD to FR system was not supportable. The original assumption was that the weakest point on the FD model would also be the weakest point on the FR model. After the testing stage was completed it was noted that the trying to alter the decision boundary of the FD model did not necessarily have the same effect on the FR model. For an attacker to gain more success from the LZP attack the ideal LPO system should be built on the FR system in future. Despite this downfall the success of such a minimal attack on an FR system is promising.

Looking to the future of the LZP attack there are several features that could be improved or added to produce more accurate results:
- Adding a function to the LPO that overlays the *'Initial'* image (reduced transparency) on the screen when the attacker is setting up to take the *'Final'* image. This will help keep the consistency between the two images allowing the attacker to align themselves accurately.
- White-Box testing on the FR system. This would be beneficial in observing how the results compare to the Black-Box testing. If the attacker wished to implement this in the real world, they would need to have access to the model used by the FR system being exploited. The current Black-Box version of the LZP is more universally testable against unknown FR systems.
- The current attack is difficult to setup as the LZP is manually pointed and aligned on the face. It would be ideal to automate this using a microcontroller that is fed the LPOs recommendation. This would speed up the testing process and increase the accuracy.
- Implementing a targeted attack feature for the LPO to test the minimum number of LZPs needed to misclassify the attacker as another individual. The idea is to use the LZP to push the attackers face across the decision boundary of a different classification class (another individual). This would require the White-Box implementation.

Although the initial IRH attack was flawed the LZP attack was a successful way to create an adversarial attack on facial recognition. For an attacker to utilize the LZP in its optimal form it is advised that the LPO be built on the system that the attacker is intending to exploit. The untargeted nature of the attack in this paper proved highly effective in decreasing the confidence score of identifying an attacker. This research plays an important role in exposing the weaknesses of DNN's. The more we try and exploit these systems the more other researchers can explore prevention techniques for the future.